\documentclass[letterpaper]{article} 
\usepackage[]{aaai24}  
\usepackage{times}  
\usepackage{helvet}  
\usepackage{courier}  
\usepackage[hyphens]{url}  
\usepackage{graphicx} 
\usepackage{natbib}  
\usepackage{caption} 
\usepackage{color}
\usepackage{soul}
\usepackage{ragged2e} 
\usepackage{bbding} 
\usepackage{booktabs,makecell, multirow, tabularx}
\usepackage{booktabs}
\usepackage{algorithm}
\usepackage{algorithmicx}
\usepackage{algpseudocode}
\usepackage{amsmath}
\usepackage{amssymb}
\usepackage{tablefootnote}
\usepackage{newfloat}
\usepackage{listings}
\DeclareCaptionStyle{ruled}{labelfont=normalfont,labelsep=colon,strut=off} 
\floatstyle{ruled}
\newfloat{listing}{tb}{lst}{}
\floatname{listing}{Listing}
\title{Adapting Pre-trained Language Models to Vision-Language Tasks \\ via Dynamic Visual Prompting}

\author {
    Shubin Huang\textsuperscript{\rm 1},
    Qiong Wu\textsuperscript{\rm 1},
    Yiyi Zhou\textsuperscript{\rm 12\thanks{corresponding author}},
    Weijie Chen\textsuperscript{\rm 3},
    Rongsheng Zhang\textsuperscript{\rm 3},
    Xiaoshuai Sun\textsuperscript{\rm 12},
}
\affiliations {
    \textsuperscript{\rm 1} Key Laboratory of Multimedia Trusted Perception and Efficient Computing,\\ Ministry of Education of China, Xiamen University, 361005, P.R. China.\\
    \textsuperscript{\rm 2}
Institute of Artificial Intelligence, Xiamen University, 361005, P.R. China.\\
    \textsuperscript{\rm 3} Fuxi AI Lab, NetEase Inc., China \\
    \{shubinhuang, qiong\}@stu.xmu.edu.cn, \{zhouyiyi, xssun\}@xmu.edu.cn, \\ \{chenweijie05, zhangrongsheng\}@corp.netease.com
}

\usepackage{bibentry}

\begin{document}
\maketitle

\begin{abstract}
Pre-trained language models (PLMs) have played an increasing role in vision-language (VL) learning, but they usually require a deep multi-modal branch for VL reasoning, resulting in excessive computation and memory overhead. Recently, visual prompting is a feasible way to adapt PLMs to VL tasks, but we notice that the use of all visual tokens will greatly execrate the already high computation, and the token placement is also vital to performance. Based on these observations, we propose a novel transfer learning approach for PLMs in this paper, termed \emph{\textbf{D}ynamic \textbf{V}isual \textbf{P}rompting} (DVP). Concretely, DVP first deploys a cross-attention module to obtain text-related and compact visual prompt tokens, thereby greatly reducing the input length of PLMs. To obtain the optimal placement, we also equip DVP with a reinforcement-learning based search algorithm, which can automatically merge DVP with PLMs for different VL tasks via a very short search process. In addition, we also combine DVP with the recently popular adapter approach to keep the most parameters of PLMs intact during adaption, which also help PLMs achieve a quick shift between single- and multi-modal tasks. We apply DVP to two representative PLMs, namely BERT and T5, and a recent large language model called LLaMA. Extensive experiments are conducted on a set of VL reasoning benchmarks including VQA2.0, GQA, SNLI-VE and ScienceQA. The experimental results not only show the merits of DVP in performance and efficiency, \emph{e.g.} +2.28\% accuracy and -80\% FLOPs on VQA2.0, but also confirm its superiority in adapting pre-trained language models to VL tasks. Our code is anonymously released at \textcolor{cyan}{\url{https://github.com/hsb1357173526/Dynamic_Visual_Prompting}}.
\end{abstract}

\section{Introduction}
Recent years have witnessed the rapid development of pre-trained language models (PLMs) \cite{devlin2018bert, lewis2019bart, liu2019roberta, brown2020language, raffel2020exploring}, which have become the \emph{de facto} standard in natural language processing (NLP). Recently, the emergence of ChatGPT \cite{chatgpt} and LLaMA \cite{touvron2023llama} further confirms the significance of PLMs in exploring \emph{general artificial intelligence}. The advent of PLMs also leads to the prevalence of large-scale pre-training in the vision-language (VL) field \cite{lu2019vilbert, tan2019lxmert, li2020oscar, shen2021much,dou2022empirical}.

However, directly adapting PLMs to VL tasks is prohibitively expensive. Above all, PLMs often act as a building block for advanced VL models. Similar to vision models \cite{ren2015faster, he2016deep, dosovitskiy2020image}, PLMs are only used as a  modal-specific encoder. For VL reasoning tasks like Visual Question Answering (VQA) \cite{goyal2017making}, the VL models still need to employ a deep fusion branch upon the multi-modal encoders, making the model extremely cumbersome. In addition to parameter redundancy, this deployment also undermines the ability of PLM in context reasoning since it only serves to embed the text words. To this end, the VL model often requires another large-scale pre-training on massive VL data \cite{lu2019vilbert,tan2019lxmert,chen2020uniter,li2020oscar,kim2021vilt,dou2022empirical}.

\begin{figure}
\centering
\includegraphics[width=0.95\columnwidth,height=0.28\textwidth]{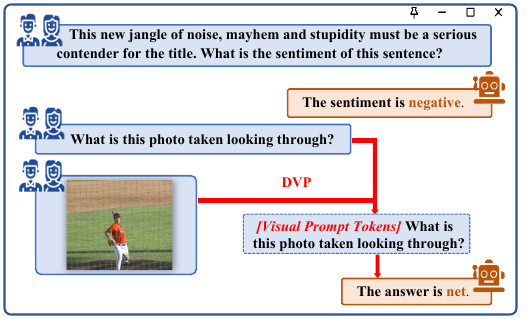}
\caption{The application of the proposed \emph{\textbf{D}ynamic \textbf{V}isual \textbf{P}rompting} (DVP) to pre-trained language models (PLMs). With low parameter and training expenditure, DVP can adapt PLMs to vision-language tasks and help them make a quick shift between single- and multi-modal tasks. }
\label{fig1}
\end{figure}

In this case, we aim to directly adapt PLMs to vision-language tasks with low parameter and training overhead. A natural solution is to directly feed all the extracted visual features into PLMs as the prompt tokens. Here, we term it \emph{common visual prompting}. This strategy has been recently attempted in some recent advances, such as FewVLM\cite{jin2021good} and BLIP-2 \cite{li2023blip}. In this paper, we also conduct some toy experiments on VL benchmarks, of which results show that this strategy can help BERT \cite{devlin2018bert} obtain decent capabilities on VL tasks, \emph{e.g.} 68.0\% on VQA2.0 \cite{goyal2017making} and 76.3\% on SNLI-VE \cite{xie2019visual}. Although the performance still lags behind large-scale vision-language pre-trained (VLP) models, such as CLIP-ViL \cite{shen2021much} and METER \cite{dou2022empirical}, it yields an affordable way for quick VL adaption. 

Nevertheless, visual prompting greatly increases the computation complexity of PLMs. For instance, directly adding all the visual patches of ViT \cite{dosovitskiy2020image} to BERT \cite{devlin2018bert} will increase the already high computation by up to 384\%. Moreover, the use of complete image features is often redundant to the VL models, as revealed in the previous VL research \cite{yang2016stacked, zhou2019dynamic, zhou2021trar}. During experiments, we also observe that the deployment of visual prompts greatly affects the final performance, \emph{i.e.} at which layer we insert the prompts. For instance, the performance of BERT ranges from 64.97\% to 69.54\% on VQA2.0 with different placements.

To address these shortcomings, we propose a novel multi-modal transfer learning approach for PLMs called \emph{Dynamic Visual Prompting} (DVP) in this paper. Instead of using the entire image features, DVP first applies a cross-attention module to collect task-related and text-relevant visual information, which is further linearly projected onto the semantic space of PLMs as the prompt token. Meanwhile, to automatically set the optimal placement of these tokens, we also equip DVP with a novel search algorithm based on $k$-armed bandit theory, which regards the token insertion as a policy action and effectively estimates the search weights via numerous single-shot trials \cite{zhou2020k}. In addition, to achieve parameter-efficient adaption of PLMs, we also combine DVP with the recently popular adapter approach \cite{sung2022vl} for downstream VL tasks. Such a combination can not only greatly save the parameter expenditure, but also enable PLMs to make a quick shift between NLP and VL tasks, as shown in Fig. \ref{fig1}. 

To validate DVP, we apply it to two representative PLMs, which are the encoder-based BERT \cite{devlin2018bert} and the encoder-decoder based T5 \cite{raffel2020exploring}. Extensive experiments are performed on a set of VL reasoning benchmarks including VQA2.0 \cite{goyal2017making}, GQA \cite{hudson2019gqa} and SNLI-VE \cite{xie2019visual}. To validate the generalization, we also apply DVP to a recent LLM called LLaMA \cite{touvron2023llama} on ScienceQA \cite{lu2022learn}. The experimental results show that compared with the common visual prompting, DVP can save up to 80\% computation while improving the performance significantly, \emph{e.g.} +1.21\% on VQA2.0 for BERT and +2.19\% on GQA for T5. When combined with Adapter \cite{sung2022vl}, DVP can well maintain the overall performance of PLMs on these benchmarks, and it only needs to update about 5.0\%-6.0\% parameters of the model for VL adaptions. In addition to pre-training costs, DVP also exhibits much better efficiency than VLP models, \emph{e.g.} 4.0G FLOPs of BERT-DVP \emph{v.s.} 956.4G FLOPs of OSCAR \cite{li2020oscar}. Meanwhile, when combined with Adapter, the updated parameters are only about 5.1\% of OSCAR. These results well confirm our motivation about DVP. 

Overall, the contributions of this paper are three-fold:
\begin{itemize}
\item We propose to directly adapt PLMs as a stand-alone model to VL tasks via inserting visual prompt tokens, which can avoid the building of heavy fusion networks and make use of the context reasoning ability of PLMs. 
\item We propose a novel transfer learning method called \emph{\textbf{D}ynamic \textbf{V}isual \textbf{P}rompting} (DVP) for efficient PLM adaption, which includes a cross-attention module to obtain compact visual tokens and a $k$-armed bandit based search algorithm for automatic prompt placement. 
\item DVP can reduce the computation of common visual prompting methods by up to 80\% while achieving better performance on multiple benchmarks. Compared with VLP models, DVP can also help PLMs obtain competitive performance on VL tasks while saving the parameters and training overhead substantially. 
\end{itemize}

\section{Related work}
With the great success of pre-training and fine-tuning paradigm in NLP, large-scale vision-language (VL) pre-training has also become a standard step in VL research \cite{chen2023vlp}. In the early stage, the pre-trained vision encoders are introduced to provide the region knowledge. VisualBERT \cite{li2019visualbert}, ViLBERT \cite{lu2019vilbert}, LXMERT \cite{tan2019lxmert}, Uniter\cite{chen2020uniter}, Oscar\cite{li2020oscar} use the pre-trained Faster-RCNN \cite{ren2015faster} to extract region features and combine them with text features for deep fusion. To break through the limitation of object detectors in inference efficiency, PixelBERT \cite{huang2020pixel}, Grid-VLP \cite{yan2021grid}, SOHO \cite{huang2021seeing} uses ResNet \cite{he2016deep} to extract the grid features of the image. These methods not only realize end-to-end training of VLP models, but also achieve better performance on VL tasks. Although VLP models achieve excellent performance on downstream VL tasks, the PLMs are only used as a language encoder and still require another large fusion branch and expensive VL pretraining. 

Recently, Prompt tuning \cite{gu2021ppt,liu2021p, wei2021finetuned,han2022ptr} aims to insert a set of prompt tokens into the input sequence of PLMs, thereby alleviating the difference between the data distributions of pre-training and downstream tasks \cite{liu2023pre}. To avoid laborious manual tuning, recent advances resort to learnable tokens for downstream tasks, which is also termed \emph{soft prompting} \cite{lester2021power}. The great success of prompt tuning in NLP also sparks its application to computer vision (CV) and vision-language (VL) studies. In terms of VL studies, CoOp\cite{zhou2022learning} keeps the parameters of CLIP unchanged, and puts learnable parameters as soft prompting for the input text, which fully uses the zero-shot retrieval ability of CLIP. To solve the problem that CoOp is easy to overfit on the basic classes, CoCoOp \cite{zhou2022conditional} implements instance-adaptive by adding the vision feature of each image to learnable prompts, which makes prompt tuning more robust. The above methods mainly use text words or learnable tokens as the prompts for NLP and VL models. There are also some very recent works applying image information for VL task prompting. With an encoder-decoder architecture, SimVLM \cite{wang2021simvlm} uses the patch features extracted from images by ViT as prefix. Frozen \cite{tsimpoukelli2021multimodal} based on GPT only trains image encoder, and uses extracted visual features as visual prompt tokens for language model, which achieves excellent performance. Overall, existing visual prompting methods often use the entire image features for VL adapting, while the critical issues of redundant visual information and excessive computation are still overlooked. 

\section{Method}
\begin{figure*}
\centering
\includegraphics[width=\textwidth]{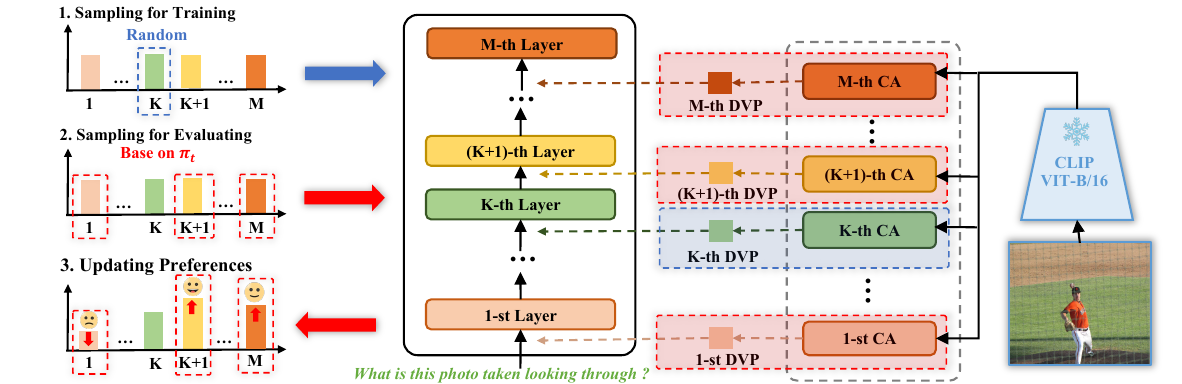}
\caption{Illustration of the proposed method. The right sub-figure illustrates the process of \emph{dynamic visual prompting} (DVP), which can produce compact yet highly related visual tokens for prompting. The left sub-figure depicts the process of the proposed \emph{k-armed bandit based automatic prompt placement} algorithm (KAB-APP). With a short search period, KAB-APP can find out the optimal deployment of DVP on different PLMs for different VL tasks.}
\label{fig2}
\end{figure*}

\subsection{Preliminary}\label{Preliminary}

Before introducing our approach, we first recap the basic visual prompting for PLMs. Concretely, given a PLM as $G(\cdot)$, and the image-text example as $(I,T)$, the target of adapting PLMs to VL tasks is to minimize the loss defined by

\begin{equation}
\setlength{\abovedisplayskip}{3pt}
\setlength{\belowdisplayskip}{3pt}
\begin{aligned}
\operatorname*{argmin}_{\theta} \mathcal{L}\big(G(I,T|\theta)\big),
\end{aligned}
\label{normal_objective}
\end{equation}
where $\theta$ is the parameters of $G$. $\mathcal{L}$ is the objective function of the downstream task. For visual prompting, a natural solution is to use the all visual features as the prompt tokens, and then project them onto the semantic space of PLMs. 

However, in this solution, the length of input sequence $L$ is largely increased by the number of visual tokens $N$, which leads to the additional computations by $O(N^2+2NL)$. Besides, using all visual information for VL reasoning is often redundant to the model \cite{rao2021dynamicvit,ryoo2021tokenlearner,bolya2022token}.

To alleviate the computational burdens of PLMs and reduce the redundancy in visual prompts, we propose to use a lightweight network $\phi$ to generate compact visual prompts, thereby reducing the length of input sequence. The optimization objective can be further defined as

\begin{equation}
\setlength{\abovedisplayskip}{3pt}
\setlength{\belowdisplayskip}{3pt}
\begin{aligned}
\operatorname*{argmin}_{\theta,\phi} \mathcal{L}\big(G(\phi(I,T),T|\theta)\big),
\end{aligned}
\label{our_objective}
\end{equation}
which can obtain substantial improvements in both model inference and computation overhead.

\subsection{Dynamic Visual Prompting}\label{Dynamic visual prompting}
In this paper, we propose a \emph{Dynamic Visual Prompting} (DVP) approach towards efficient VL adaption. In principle, DVP adopts cross-modal attention to dynamically generate visual prompts relevant to the input text and image, as shown in the right half of Fig. \ref{fig2}.

Given an image $I$, we first use a frozen visual encoder to extract its features, denoted as $\mathbf{F}_v \in \mathbb{R}^{N \times d}$, where $N$ is the number of visual features and $d$ is the feature dimension. To obtain the compact visual tokens related to the input text, we use the text features in PLMs as the \emph{query} vectors, $\mathbf{F}_t \in \mathbb{R}^{L \times d}$, where $L$ refers to the length of text sentences. With the projection weight matrices, $\mathbf{W}_Q \in \mathbb{R}^{d \times d}$, $\mathbf{W}_K \in \mathbb{R}^{d \times d}$, and $\mathbf{W}_V \in \mathbb{R}^{d \times d}$, for $Q, K, V$ transformations in our cross-modal attention module $\phi$. DVP is obtained by

\begin{equation}
\setlength\abovedisplayskip{6pt}
\begin{aligned}
\mathbf{F}_{DVP} = Softmax(\frac{\mathbf{F}_t\mathbf{W}_Q(\mathbf{F}_v\mathbf{W}_K)^T}{\sqrt{d}})\mathbf{F}_v\mathbf{W}_V.
\end{aligned}
\label{cross_attn}
\end{equation}

Here, we also employ multi-head attention mechanism, so the above formula can be further rewritten as

\begin{equation}
\begin{aligned}
\mathbf{F}_{DVP} &= Concat(head_1, ..., head_h)\mathbf{W}_O, \\
head_i &= Softmax(\frac{\mathbf{F}_t\mathbf{W}^i_Q(\mathbf{F}_v\mathbf{W}^i_K)^T}{\sqrt{\frac{d}{n}}})\mathbf{F}_v\mathbf{W}^i_V,
\end{aligned}
\label{multi_head_cross_attn}
\end{equation}

where $n$ is the number of attention heads, $\mathbf{W}_O \in \mathbb{R}^{d \times d}$ and $\mathbf{W}^i_Q, \mathbf{W}^i_K, \mathbf{W}^i_V \in \mathbb{R}^{d \times \frac{d}{n}}$ are the weight matrices. To this end, the length of generated $\mathbf{F}_{DVP}$ becomes $L$, and the computational complexity is reduced from $O\big((N+L)^2\big)$ to $O\big((L+L)^2\big)$ in calculating attention matrix, where $N$ is often larger than $L$ in VL tasks \cite{kim2021vilt,shen2021much,dou2022empirical}. 

However, the computation overhead is still expensive and we further reduce the length of prompt tokens to improve the efficiency by using the global features of PLMs as \emph{query} vectors, such as \emph{[CLS]} token of BERT, pooling feature of T5's encoder or \emph{[EOS]} token of LLaMA.

For encoder-based language models like BERT, we define its layers as $\Theta=[\Theta_1,...,\Theta_{K},...,\Theta_{M}]$, where $K$ is the index of insertion layer and $M$ refers to the number of layers. Here, $\mathbf{F}_t$ denotes the hidden features obtained before the $K$-$th$ layer we want to insert, and we use its \emph{[CLS]} token features $\mathbf{F}_t^{[CLS]}$ as \emph{query} vector. Afterwards, DVP of BERT at insertion layer $K$ is obtained by
\begin{equation}
\setlength{\abovedisplayskip}{1pt}
\setlength{\belowdisplayskip}{1pt}
\begin{aligned}
\mathbf{F}_t &= \Theta_{K-1}\big(\Theta_{K-2}...\Theta_1\big(E(T)\big)\big), \\
\mathbf{F}_{DVP} &= \phi(\emph{query}=\mathbf{F}_t^{[CLS]}, \emph{key}=\mathbf{F}_v, \emph{value}=\mathbf{F}_v),
\end{aligned}
\label{eq1}
\end{equation}
where $E$ denotes the text embedding. Next, $\mathbf{F}_{DVP}$ is concatenated with $\mathbf{F}_t$ and sent to subsequent layers
\begin{equation}
[\mathbf{F}_{DVP};\mathbf{F}_t] = \Theta_M\big(\Theta_{M-1}...\Theta_{K}\big([\mathbf{F}_{DVP};\mathbf{F}_t]\big)\big) ,
\label{eq2}
\end{equation}
Finally, we use $\mathbf{F}_t^{[CLS]}$ to connect the specific VL task classification head for prediction.

For encoder-decoder based PLMs like T5, we define its encoder as $\Psi$, decoder layers as $\Theta=[\Theta_1,...,\Theta_{K},...,\Theta_{M}]$ and decoder input vector as $\mathbf{f} \in \mathbb{R}^{d}$. Here, $\mathbf{F}_t$ means the output of $\Psi$ and we utilize the pooling features of $\rho(\mathbf{F}_t)$ as text \emph{query} vector, where $\rho$ means the operation of average mean pooling. Then $\mathbf{F_{DVP}}$ of T5 at insertion decoder layer $K$ is obtained by
\begin{equation}
\begin{aligned}
\mathbf{F}_t &= \Psi\big(E(T)\big), \\
\mathbf{F}_{DVP} &= \phi(\emph{query}=\rho(\mathbf{F}_t), \emph{key}=\mathbf{F}_v, \emph{value}=\mathbf{F}_v).
\end{aligned}
\label{eq3}
\end{equation}
Afterwards, $\Theta_K$'s output is spliced with $\mathbf{F}_{DVP}$ and sent to the subsequent decoder layers:
\begin{equation}
\setlength{\abovedisplayskip}{1pt}
\setlength{\belowdisplayskip}{1pt}
\begin{aligned}
\mathbf{f} &= \Theta_{K-1}\big(\Theta_{K-2}...\Theta_1\big(E(f)\big)\big), \\
[\mathbf{F}_{DVP};\mathbf{f}] &= \Theta_M\big(\Theta_{M-1}...\Theta_{K}\big([\mathbf{F}_{DVP};\mathbf{f}]\big)\big),
\end{aligned}
\label{eq4}
\end{equation}
Lastly, we use $\mathbf{f}$ attached to the classification head for prediction.

\subsection{Automatic Prompt Placement}\label{Auto search}
In our experimental trials, we find that the model performance is disparate with the different placements of DVP. Meanwhile, for VL tasks with various definitions, a static setting is suboptimal to PLM adaption. To this end, we propose a \textit{\textbf{$k$}-\textbf{A}rmed \textbf{B}andit based \textbf{A}utomatic \textbf{P}rompt \textbf{P}lacement} algorithm (KAB-APP) as shown in the left half of Fig. \ref{fig2}.

\textbf{Problem Definition.} The target of KAB-APP is to automatically find out the insertion layer that can lead to the best performance of DVP. Thus, we consider the prompt placement as a policy action and model the search process as a $k$-armed bandit problem.

Concretely, we equally initialize the action preference for each layer of PLMs as $h \in \mathbb{R}$, that is, the probability of each layer being selected. During each search step, the policy for the $K$-$th$ layer $\pi_t(K)$ is obtained by
\begin{equation}
\begin{aligned}
\pi_t(K) &= \frac{e^{\mathbf{H}_t(K)}}{\sum_{m=1}^{M}e^{\mathbf{H}_t(l)}},
\end{aligned}
\label{action}
\end{equation}
where $\mathbf{H}_t=[h_t^1,...,h_t^K,...,h_t^M]$ denotes the preference of each layer at $t$-th setp. Based on Eq. \ref{action}, the action weight is then updated by
\begin{equation}
\begin{aligned}
\Delta R &= R_t-R_b, \\
\mathbf{H}_{t+1}(K) &= \mathbf{H}_{t}(K)+\alpha\Delta R\pi_t(K)(1-\pi_t(K)), 
\end{aligned}
\label{eq5}
\end{equation}
where $\alpha$ is the learning rate for updating $\mathbf{H}$, $R_t$ is the reward of inserting DVP to $K$-$th$ layer at the $t$-$th$ step, $R_b$ is the baseline reward obtained by $n$ sampling times. Here we use the average value of $n$ samplings' rewards at the $t$-$th$ step. 

\begin{algorithm}[h]
  \caption{KAB-APP} 
  \begin{algorithmic}[1]
    \Require
      The training and val sets $D_t$ and $D_v$. Model Layers 
      \Statex $\qquad \Theta = [\Theta_1, ..., \Theta_{M}]$.
    \Ensure
       The optimal insertion layer $K$.
       \State Assign a cross-modal attention layer to each insertion layer and initialize their weights $\Phi = [\phi_1, ..., \phi_{M}]$.
       \State Initialize the preferences of all layers, $\mathbf{H}_0 = [h_0^1, ..., h_0^{M}]$.
        \For{$t$ in $T$ Steps}
        \State Perform sampling from all possible insertion layers.
        \State Obtain sampled insertion layer $K$ and corresponding $\phi_K$.
         \State Obtain the sampled insertion layer's DVP by Eq.{\ref{eq1}} or Eq.{\ref{eq3}}.
         \State Update weights of $\Theta, \phi_K$ by $\mathcal L_{train}$.
        \State Sample $n$ layers in $[1,...,M]$ via $H_{t}$ by Eq.\ref{action}.
        \State Obtain a val batch $d_{v}^{t} \xleftarrow{} D_v$.
        \For{$i$ in $n$ insertion layers}
        \State Obtain $R_t^{i} \xleftarrow{} Pred(\Theta, \phi_i, d_v^t)$.
        \EndFor
        \State Obtain baseline reward $R_b = \frac{1}{n} \Sigma_{i=1}^n R^{i}_{t}$.
        \For{$i$ in $n$ insertion layers}
        \State Update $\mathbf{H}_t$ based on $R_t^i, R_b$ by Eq.\ref{eq5}.
        \EndFor
      \EndFor
    \State Obtain the optimal insert layer $K$  based on $\mathbf{H}_T$.
    \State \Return $K$
  \end{algorithmic}
\end{algorithm}

\textbf{Search Algorithm.} The procedure of KAB-APP is described in \textit{Algorithm 1}. Firstly, we establish an action space consisted of $M$ insertion layers for PLMs. Meanwhile, we equip all the possible insertion layers with independent cross-modal attention modules.

At each training step, an inserted layer is randomly sampled from $M$, and this layer $K$ and the corresponding cross-attention module $\phi_K$ are activated for training. In a short training interval, we sample $n$ candidate insertion layers for validation by the RL policy $\pi_t$, and use the validation accuracy or the power of the negative loss of $e$, as the rewards. Then we update preferences $\mathbf{H}$ by using the gradient update rules in Eq. \ref{eq5}.

KAB-APP aims to search the best insertion layer for DVP on the language model. With the above search algorithm, the action weights can be fairly estimated via numerous \emph{single-shot} validations \cite{zhou2020k}. Since the search space is limited, the whole process can be accomplished within a few periods. In practice, we control this search process to the limited steps, which only requires an additional 30-40\% of the time of fine-tuning a PLM.

\subsection{Combination with Adapter}\label{Combination with adapter}

We further combine DVP with an advanced parameter-efficient transfer learning approach called Adapter \cite{sung2022vl,luo2023cheap} to reduce the scale of training parameters. It can also help PLMs to make a quick shift between NLP and VL tasks. In practice, given a PLM, we can search the optimal DVP insertion layer $K$ with KAB-APP, and then use cross-modal attention layer $\phi$ to generate lightweight DVP at $K$-th layer. Furthermore, we equipped language model with Adapters. The process above not only satisfies high parameter efficiency, but also attains optimal performance.

\section{Experiments}
\subsection{Datasets}
\textbf{VQA2.0} \cite{goyal2017making} is a benchmark dataset that builds on VQA1.0 \cite{antol2015vqa} and contains open-ended questions about images from MSCOCO \cite{ren2015exploring}. Following previous works \cite{yu2019deep}, we consider VQA2.0 as a classification task with 3129 answer categories. Following ViLT \cite{kim2021vilt}, we use VQA2.0 \emph{train} set and \emph{val} set for fine-tuning while retaining 1000 examples of the \emph{val} set for validation.

\textbf{GQA} \cite{hudson2019gqa} is a challenging benchmark for visual reasoning. GQA contains 22$M$ questions and 113$K$ images from Visual Genome \cite{krishna2017visual}. Similar with VQA2.0, we design GQA as a classification task with 1843 answer classes. We use the \emph{train} set for training and then report the performance on \emph{test-dev} set.

\textbf{SNLI-VE} \cite{xie2019visual} is a dataset proposed for the visual entailment task and build upon SNLI \cite{bowman2015large} and Flickr30K \cite{plummer2015flickr30k}. This task is to predict the relationship between the given image and sentence from three possible relations, \emph{i.e.} \emph{entailment}, \emph{neutral}, and \emph{contradiction}. We use the \emph{train} set for training and then report the performance on \emph{dev} set and \emph{test} set.

\textbf{ScienceQA} \cite{lu2022learn} is a comprehensive multimodal dataset tailored for science question answering. It encompasses 21,208 questions, drawing from 3 subjects and spanning 26 topics and 127 categories. Notably, the dataset is split into text-only and text-image examples, segmented as \emph{train} set (12,726 examples), \emph{val} set (4,241 examples), and \emph{test} set (4,241 examples). We use the \emph{train} set for fine-tuning and evaluate performance on the \emph{test} set.

\textbf{Metrics}. For VQA2.0, we apply the VQA accuracy \cite{antol2015vqa} as the metric. Thus, for GQA and SNLI-VE, we use the classification accuracy. As for ScienceQA, we evaluate performance using average accuracy.

\begin{table}[]
\caption{Comparison with other visual prompting solutions. ``\emph{[CLS]} prompting'' denotes the use of the \emph{[CLS]} token of the image encoder as the prompt. The suffixes of DVP indicates that single or multiple dynamic tokens are used. Note that all tokens are inserted at the 1-$st$ layer.}
\label{linear_vs_dvp}
\renewcommand\arraystretch{0.8}
\setlength{\tabcolsep}{1.5mm}{
\begin{tabular}{@{}cccccc@{}}
\toprule
\textbf{Method}                                                                 & \textbf{Model} & \textbf{FLOPs} & \textbf{\begin{tabular}[c]{@{}c@{}}VQAv2\end{tabular}} & \textbf{\begin{tabular}[c]{@{}c@{}}GQA\end{tabular}} & \textbf{\begin{tabular}[c]{@{}c@{}}SNLI-VE\end{tabular}} \\ \midrule
\multirow{2}{*}{\begin{tabular}[c]{@{}c@{}}Common \\ Prompting\end{tabular}}    & BERT           & 167.6G         & 68.00                                                             & 51.34                                                           & 76.26                                                                   \\
                                                                                & T5             & 192.4G         & 69.24                                                             & 50.85                                                           & 76.19                                                                   \\ \midrule
\multirow{2}{*}{\begin{tabular}[c]{@{}c@{}}{[}CLS{]} \\ Prompting\end{tabular}} & BERT           & 34.2G          & 65.50                                                             & 50.87                                                           & \textbf{77.28}                                                          \\
                                                                                & T5             & 36.8G          & 65.17                                                             & 48.89                                                           &  76.70                                                             \\ \midrule
\multirow{2}{*}{DVP$_{multi}$}                                                     & BERT           & 44.8G          & 66.35                                                             & 50.22                                                           & 75.95                                                                   \\
                                                                                & T5             & 49.2G          &69.30                                                       &51.53                                                     & 76.50                                                                   \\ \midrule
\multirow{2}{*}{DVP$_{single}$}                                                            & BERT           & 34.6G          & 64.97                                                             & 50.01                                                           & 75.78                                                                   \\
                                                                                & T5             & 37.2G          & \textbf{69.42}                                                    & \textbf{52.27}                                                  & 75.60                                                                   \\ \bottomrule
\end{tabular}}
\end{table}

\subsection{Experimental Settings}
For vision encoder, we choose CLIP-ViT-B/16 or CLIP-ViT-L/14 \cite{radford2021learning} to extract image features. The resolution of input images is resized to 224 $\times$ 224 and the vision encoder keeps frozen. We use AdamW optimizer \cite{loshchilov2017decoupled} with learning rates of $1 \times 10^{-4}$ , $2 \times 10^{-4}$ for BERT-base and T5-base, respectively. In ScienceQA, we apply AdamW optimizer with base learning rate of $1 \times 10^{-2}$ for LLaMA-7B. The number of insertion layers per search samplings $n$ is set to 5 for BERT-base and T5-base, and to 10 for LLaMA-7B. In terms of KAB-APP, the search epochs of KAB-APP are set to 2, 2, 1 and 2 for VQA2.0, GQA, SNLI-VE and ScienceQA, respectively. The learning speed $\alpha$ of KAB-APP is set to $5 \times 10^{-3}$ for BERT-base and T5-base, and to 1 for LLaMA-7B. More experimental settings are given in the appendix.

\subsection{Experimental Results}

\begin{table}[]
\caption{The results of different prompt placement of DVP on PLMs. Here, we manually set the placement at 1-$st$, 4-$th$, 7-$th$, 10-$th$ and 12-$th$ layers of two PLMs.}
\label{insert_dvp}
\renewcommand\arraystretch{0.8}
\setlength{\tabcolsep}{3mm}{
\begin{tabular}{@{}ccccc@{}}
\toprule
\textbf{Model}            & \textbf{\begin{tabular}[c]{@{}c@{}} Layer\end{tabular}} & \textbf{\begin{tabular}[c]{@{}c@{}}VQAv2\end{tabular}} & \textbf{\begin{tabular}[c]{@{}c@{}}GQA\end{tabular}} & \textbf{\begin{tabular}[c]{@{}c@{}}SNLI-VE\end{tabular}} \\ \midrule
\multirow{5}{*}{BERT-DVP} & 1-$st$                                                              & 64.97                                                             & 50.01                                                           & \textbf{75.78}                                                          \\
                          & 4-$th$                                                              & 68.22                                                             & 50.91                                                           & 75.11                                                                   \\
                          & 7-$th$                                                              & 68.92                                                             & 52.07                                                           & 74.26                                                                   \\
                          & 10-$th$                                                             & \textbf{69.54}                                                    & 52.18                                                           & 74.76                                                                   \\
                          & 12-$th$                                                             & 68.76                                                             & \textbf{52.55}                                                  & 73.72                                                                   \\ \midrule
\multirow{5}{*}{T5-DVP}   & 1-$st$                                                              & 69.42                                                             & 52.27                                                           & \textbf{75.60}                                                          \\
                          & 4-$th$                                                              & 69.15                                                             & 51.86                                                           & 75.04                                                                   \\
                          & 7-$th$                                                              & 69.67                                                             & 52.07                                                           & 74.97                                                                   \\
                          & 10-$th$                                                             & \textbf{69.75}                                                    & \textbf{53.04}                                                  & 74.81                                                                   \\
                          & 12-$th$                                                             & 69.37                                                             & 52.15                                                           & 75.04                                                                   \\ \bottomrule
\end{tabular}}
\end{table}

\begin{table}[]
\caption{Comparison between KAB-APP and the Manual search. ``Com-Prompt.'' refers to the common visual prompting mentioned above. ``At 1$st$-layer'' denote that deploying DVP at the input layer of PLMs.}
\label{kab_app_vs}
\renewcommand\arraystretch{0.8}
\setlength{\tabcolsep}{1mm}{
\begin{tabular}{@{}cccccc@{}}
\toprule
\textbf{Model}        & \textbf{Setting} & \textbf{FLOPs} & \textbf{\begin{tabular}[c]{@{}c@{}}VQAv2\end{tabular}} & \textbf{\begin{tabular}[c]{@{}c@{}}GQA\end{tabular}} & \textbf{\begin{tabular}[c]{@{}c@{}}SNLI-VE\end{tabular}} \\ \midrule
\multirow{4}{*}{BERT} & Com-Prompt.      & 167.6G         & 68.00                                                             & 51.34                                                           & \textbf{76.26}                                                          \\
                      & At 1$st$-layer    & 34.6G          & 64.97                                                             & 50.01                                                           & 75.78                                                                   \\
                      & Manual           & 34.1G          & \textbf{69.54}                                                    & 52.55                                                           & 76.00                                                                   \\
                      & KAB-APP          & 34.0G          & 69.21                                                             & \textbf{52.55}                                                  & 74.76                                                                   \\ \midrule
\multirow{4}{*}{T5}   & Com-Prompt.      & 192.4G         & 69.24                                                             & 50.85                                                           & \textbf{76.19}                                                          \\
                      & At 1$st$-layer \tablefootnote{The first layer of the decoder of T5.}    & 37.2G          & 69.42                                                             & 52.27                                                           & 75.60                                                                   \\
                      & Manual           & 36.8G          & 69.82                                                             & 53.04                                                           & 75.60                                                                   \\
                      & KAB-APP          & 36.8G          & \textbf{69.82}                                                    & \textbf{53.04}                                                  & 75.60                                                                   \\ \bottomrule
\end{tabular}}
\end{table}

\subsubsection{Quantitative Analysis}
\ 
\newline
\indent \textbf{The comparison with common visual prompting.} To validate the proposed \emph{dynamic visual prompting} (DVP), we first compare it with the common visual prompting solutions in Tab. \ref{linear_vs_dvp}, \emph{i.e.} using all image features or the \emph{[CLS]} token of the image encoder as the prompt tokens. For a fair comparison, all these methods place these tokens at the 1$st$ layer of BERT and the decoder of T5. We can first observe that using all image features is a reliable way of adapting PLMs to VL tasks, of which performance is consistent across tasks and models. However, compared with other solutions, its computation is much more expensive, \emph{i.e.} +4-+5 times. For \emph{[CLS]} prompting, using a static visual prompt is inferior for most VL tasks, especially the ones that require fine-grained reasoning, \emph{i.e.} VQA2.0 and GQA. For the DVP methods, the dynamic prompt tokens can obtain better performance than the static \emph{[CLS]} prompting in most cases, confirming our assumption about dynamic prompting. However, the increase of dynamic tokens does not always lead to better performance under all settings. Meanwhile, the effectivenesses of DVP on BERT and T5 are also different. Compared to T5, we directly place DVP at the input layer of BERT, which has yet to learn sufficient text semantics. This finding implies that the placement of prompt tokens is critical to DVP. Overall, these results confirm the feasibility of DVP, but it still needs great improvements in token placement.

\textbf{The impact of prompt placement.}
We further examine the impact of prompt placement via manual setting in Tab. \ref{insert_dvp}. From this table, we can first confirm that the prompt placement is vital to performance. In terms of BERT, the performance of different DVP placements varies vastly, which changes are 4.57\%, 2.54\%, and 2.06\% in VQA2.0, GQA, and SNLI-VE, respectively. In terms of T5, the performance change still exists but is less obvious, which are 0.6\%, 1.18\%, and 0.79\%, respectively. To explain, we only try the placement of T5 from the decoding layers, where the textual semantic is well-built via its encoder branch. In this case, the quality of DVP is better ensured, and the similar cases can be also found in Tab. \ref{linear_vs_dvp}. Furthermore, we can see that the effect is also different on VL tasks. For instance, VQA performs better when the prompts are inserted to the higher layers and SNLI-VE does the opposite. For DVP, we speculate that VQA firstly focuses on understanding the question itself, and then use its semantics to extract key information in the image to generate an effective prompt. In contrast, SNLI-VE may need more understanding of image semantics, and deploy it at a lower layer can better facilitate VL comprehension. Thus, we can conclude that the best interaction of DVP varies for different tasks, which substantially confirms the necessity of automatic prompt placement.

\textbf{Effectiveness of KAB-APP.} To validate KAB-APP, we compare it with the manual search in Tab. \ref{kab_app_vs}, which is obtained by training all possible insertions. It can be seen that the search results of KAB-APP is very close to that of manual search, which can be regarded as the upper-bound of DVP. On BERT, KAB-APP can achieve competitive performance against the manual search, although the best insertion layers are slightly different on VQA2.0, \emph{i.e.} 11-$th$ and 10-$th$ layers, respectively. In terms of T5, of which text semantics are well learned, the best placements of KAB-APP and manual search are consistent on all tasks, greatly showing the effectiveness of our search algorithm. From Tab. \ref{kab_app_vs}, we can also see that via finding the optimal placement, the performance of DVP can be greatly improved. For instance, KAB-APP is better than using all visual prompts (Com-Prompt.) on VQA2.0 and GQA, while reducing the computation by 80.7\%. Meanwhile, compared to the static placement, \emph{i.e.} ``at 1-$st$ layer'', the performance gains become more obvious. These results well confirm our assumption and validate the effectiveness of KAB-APP.

\textbf{The combination with Adapter.} To further improve efficiency, we introduce Adapter \cite{sung2022vl} to parameter-efficiently adapt PLMs into the downstream tasks. As shown in the bottom two rows of Tab. \ref{tab2}, the adapter-based methods achieve similar performance to fine-tuning. Specifically, the performance on BERT is slightly improved compared to the fine-tuning mode, \emph{e.g.} +1.07\%, +0.92\% and +0.94\% on VQA2.0, GQA and SNLI-VE, respectively. In terms of T5, the DVP with adapters can achieve 99.3\%, 99.8\% and 98.8\% performance of fine-tuning manner while only updating 5\% parameters. From the above experiments, we can conclude that the proposed DVP method can easily cooperate with other PETL methods for adapting PLMs in a more efficient way.

\begin{table*}[]
\caption{Comparison with the vision-language pre-trained (VLP) models. ``\emph{Total Params}'' does not include the parameters of the classification head. DVP$_{adp}$ denotes the combination of Adapter.}
\label{tab2}
 \renewcommand\arraystretch{0.9}
\setlength{\tabcolsep}{1.3mm}{
\begin{tabular}{@{}cccccccccc@{}}
\toprule
\multirow{2}{*}{\textbf{Model}} & \multirow{2}{*}{\textbf{\begin{tabular}[c]{@{}c@{}}Total\\ Params  \end{tabular}}} & \multirow{2}{*}{\textbf{\begin{tabular}[c]{@{}c@{}}Updated \\ Params\end{tabular}}} & \multirow{2}{*}{\textbf{\begin{tabular}[c]{@{}c@{}}Pre-training \\ Data\end{tabular}}} & \multirow{2}{*}{\textbf{\begin{tabular}[c]{@{}c@{}}Token\\ Number\end{tabular}}} & \multirow{2}{*}{\textbf{FLOPs}} & \textbf{VQAv2}    & \textbf{GQA}      & \multicolumn{2}{c}{\textbf{SNLI-VE}} \\
                                &                                                                                   &                                                                                     &                                                                                       &                                                                                  &                                 & \textbf{Test-Dev} & \textbf{Test-Dev} & \textbf{Dev}     & \textbf{Test-P}   \\ \midrule
VisualBERT \cite{li2019visualbert}                      & 198M                                                                              & 138M                                                                                & 0.6M                                                                                  & 164                                                                              & 425.0G                          & 70.60             & -                 & -                & -                 \\
ViLBERT \cite{lu2019vilbert}                         & 302M                                                                              & 258M                                                                                & 3.1M                                                                                  & 72                                                                               & 958.1G                          & 70.18             & -                 & -                & -                 \\
LXMERT \cite{tan2019lxmert}                          & 268M                                                                              & 223M                                                                                & 9.2M                                                                                  & 56                                                                               & 952.0G                          & 72.42             & 60.00             & -                & -                 \\
Uniter-Base \cite{chen2020uniter}                     & 183M                                                                              & 138M                                                                                & 9.0M                                                                                  & 96                                                                               & 949.9G                          & 72.70             & -                 & 78.59            & 78.28             \\
Oscar-Base \cite{li2020oscar}                     & 183M                                                                              & 138M                                                                                & 6.5M                                                                                  & 85                                                                               & 956.4G                          & 73.16             & 61.19             & -                & -                 \\
ViLT\cite{kim2021vilt}                    & 111M                                                                              & 111M                                                                                & 9.0M                                                                                  & 240                                                                              & \textbf{55.9G}                           & 70.85             & 57.44             & 76.69            & 76.74             \\
CLIP-ViL$_p$ \cite{shen2021much}                    & 159M                                                                              & 159M                                                                                & 9.2M                                                                                  & 120                                                                              & 80.8G                          & 76.48             & \textbf{61.42}    & 80.61            & 80.20             \\
METER \cite{dou2022empirical}                           & 323M                                                                              & 323M                                                                                & 9.0M                                                                                  & 627                                                                              & 242.5G                          & \textbf{77.68}    & -                 & \textbf{80.86}   & \textbf{81.19}    \\ \midrule
BERT-DVP                        & 198M                                                                              & 111M                                                                                & 0.0M                                                                                     & 17                                                                               & \textbf{34.0G}                           & 69.21             & 52.83             & 74.80            & 74.75             \\
T5-DVP                          & 311M                                                                              & 225M                                                                                & 0.0M                                                                                     & 18                                                                               & 36.8G                           & 69.82       & 53.12      & 75.46      & 75.60       \\
BERT-DVP$_{adp}$              & 201M                                                                              & 7M                                                                                  & 0.0M                                                                                     & 17                                                                               & 34.1G                           & \textbf{70.28}    & \textbf{53.75}    & \textbf{75.83}   & \textbf{75.77}    \\
T5-DVP$_{adp}$                & 320M                                                                              & 11M                                                                                 & 0.0M                                                                                     & 18                                                                               & 37.0G                           & 69.34             & 53.03             & 74.68            & 74.56             \\ \bottomrule
\end{tabular}}
\end{table*}

\begin{table*}[]
\caption{Comparison between the LLaMA with DVP and the other advanced multi-modal LLMs on ScienceQA.}
\label{sqa}
\renewcommand\arraystretch{1.2}
\setlength{\tabcolsep}{1.5mm}{
\begin{tabular}{cccccc}
\hline
\textbf{Model} & \textbf{VL Pre-training} & \textbf{Base Language Model} & \textbf{Total Params} & \textbf{Updated Params} & \textbf{ScienceQA} \\ \hline
BLIP-2 \cite{li2023blip}         & \Checkmark                              & Vicuna-7B                    & 7B                    & 188M                    & 77.30              \\
BLIP-2  \cite{li2023blip}       & \Checkmark                              & FlanT5$_{XXL}$               & 12B                   & 188M                    & 89.50              \\
LLaVA  \cite{liu2023visual}        & \Checkmark                              & LLaMA-13B                    & 13B                    & 13B                     & \textbf{90.92}     \\
LLaMA-DVP$_{adp}$      & \XSolid                              & LLaMA-7B                     & 7B                    & 9M                      & 89.82              \\ \hline
\end{tabular}}
\end{table*}

\textbf{Comparison with visual-language pre-trained models.} We also compare DVP with a bunch of VLP models on VL benchmarks in Tab. \ref{tab2}, which are often built with another deep fusion networks. Here, DVP is deployed according to the results of KAB-APP. Notably, compared to VLP models, DVP directly transfers PLMs into VL tasks without additional requirement of large-scale VL pre-training. When adapting pre-trained model into downstream tasks, DVP takes fewer computing resources especially when being combined with the adapters, \emph{i.e.} updates only 5.9\% and 5.0\% parameters of BERT and T5, respectively. When training similar parameters, the proposed DVP on BERT achieves competitive performance against ViLT on VQA2.0, GQA and SNLI-VE, respectively, while only taking 60.8\% FLOPs. Compare to the SOTA method METER \cite{dou2022empirical}, the proposed BERT-DVP$_{adp}$ method takes 2.2\% trainable parameters and 14.1\% FLOPs to achieve 70.28\% performance on VQA2.0. Overall, these results confirm the feasibility of DVP in adapting PLMs to VL tasks.

\textbf{The generalization ability of DVP.} In Tab.\ref{sqa}, we examine the generalization of DVP to the recently proposed LLM called LLaMA \cite{touvron2023llama} on ScienceQA. When applying DVP to LLaMA-7B, another VL pre-training is also not required, and updated parameters are only 9M, which is about 0.13\% of LLaMA-7B. Compared with BLIP-2\cite{li2023blip} using Vicuna-7B as the base language model, LLaMA-DVP$_{adp}$ updates only 4.8\% of its trainable parameters, and the performance can exceed 12.52\%. Even when BLIP-2 \cite{li2023blip} changes the base language model to FlanT5$_{XXL}$ of 12B parameters, LLaMA-DVP$_{adp}$ still has comparable performance without VL pre-training. Compared with the current SOTA method LLaVA \cite{liu2023visual} using LLaMA-13B as base language model, LLaMA-DVP$_{adp}$ can achieve 98.8\% of its performance while only updating less than 0.1\% of its parameters. These results well validate the generalization of DVP on LLM.

\subsubsection{Qualitative Analysis}
\begin{figure}
\centering
\includegraphics[width=\columnwidth]{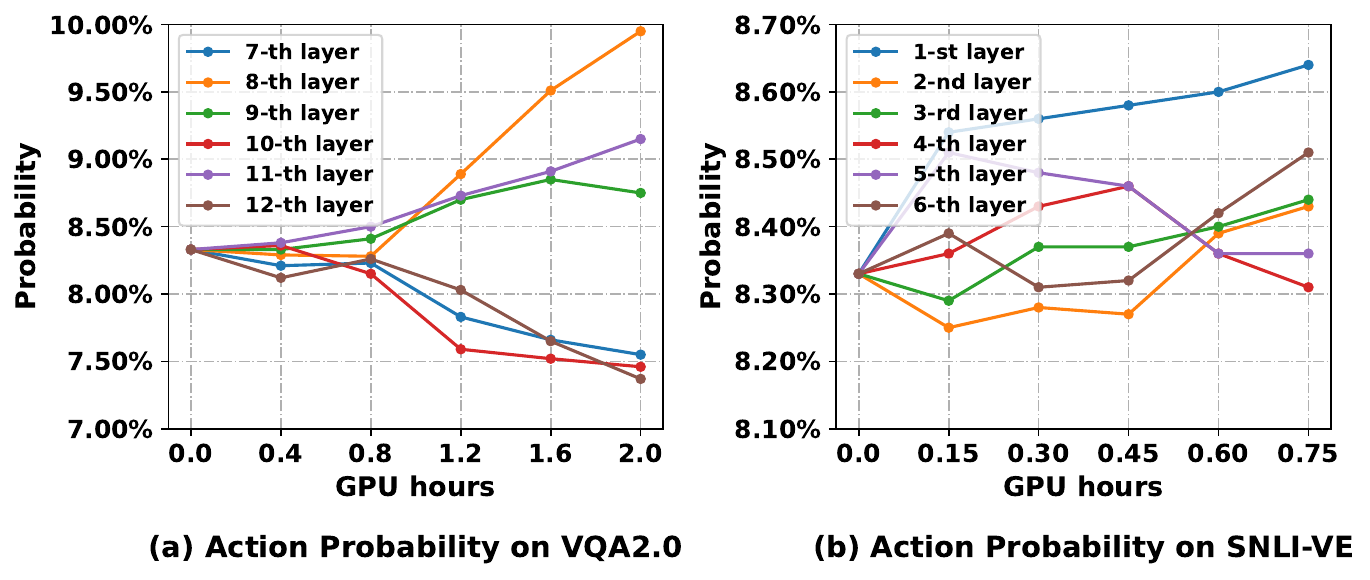}
\caption{The change of action probabilities of KAB-APP on BERT. The network search is performed on one NVIDIA-A100, which requires about 2 hours and 0.75 hours for VQA2.0 and SNLI-VE, respectively. For clarity, we depict the curves of the top-6 layers of BERT.}
\label{fig5}
\end{figure}

\ 
\newline
\indent To obtain further insight into KAB-APP, we also visualize its search process in Fig. \ref{fig5}, which illustrates the probability variations of the search layers. We can observe that as KAB-APP updates the preference of each layer, the probabilities of the optimal insertion gradually increase. In the initial stage, the preferences of possible placements are similar. After a very short period, the optimal insertion layer stands out. Overall, these search processes suggest that KAB-APP can quickly converge to the optimal placement of DVP on different PLMs for different VL tasks, of which expenditure is much cheaper than the manual search. 

\section{Conclusion}

In this paper, we focus on exploring pre-trained language models (PLMs) as a stand-alone model for VL reasoning tasks. The recently popular visual prompting is a feasible solution for this target, but it exhibits obvious information redundancy and excessive computation overhead. Motivated by this observation, we propose a novel transfer learning method to adapt PLMs on VL task, termed \textit{\textbf{D}ynamic \textbf{V}isual \textbf{P}rompt} (DVP), which obtains text-related and compact visual prompts through a cross-modal attention layer. Meanwhile, we also equip DVP with a reinforcement-learning-based search algorithm to automatically find the optimal placement, termed \textit{\textbf{k}-\textbf{A}rmed \textbf{B}andit based \textbf{A}utomatic \textbf{P}rompt \textbf{P}lacement} (KAB-APP). Furthermore, we combine DVP with the recently popular \emph{Adapter} approach to reduce the scale of updated parameters on VL tasks. The extensive experiments not only show the superior efficiency and competitive performance of DVP against other prompting solutions and existing VLP models, but also confirm that DVP is an affordable way for the quick PLM adaption. 

\section{Acknowledgements}
This work was supported by National Key R\&D Program of China (No.2022ZD0118202), the National Science Fund for Distinguished Young Scholars (No.62025603), the National Natural Science Foundation of China (No.U21B2037, No.U22B2051, No.62176222, No.62176223, No.62176226, No.62072386, No.62072387, No.62072389, No.62002305 and No.62272401), and the Natural Science Foundation of Fujian Province of China (No.2021J01002, No.2022J06001).

\bibliography{aaai24}

\begin{thebibliography}{56}
\providecommand{\natexlab}[1]{#1}

\bibitem[{Antol et~al.(2015)Antol, Agrawal, Lu, Mitchell, Batra, Zitnick, and
  Parikh}]{antol2015vqa}
Antol, S.; Agrawal, A.; Lu, J.; Mitchell, M.; Batra, D.; Zitnick, C.~L.; and
  Parikh, D. 2015.
\newblock Vqa: Visual question answering.
\newblock In \emph{Proceedings of the IEEE international conference on computer
  vision}, 2425--2433.

\bibitem[{Bolya et~al.(2022)Bolya, Fu, Dai, Zhang, Feichtenhofer, and
  Hoffman}]{bolya2022token}
Bolya, D.; Fu, C.-Y.; Dai, X.; Zhang, P.; Feichtenhofer, C.; and Hoffman, J.
  2022.
\newblock Token Merging: Your ViT But Faster.
\newblock \emph{arXiv preprint arXiv:2210.09461}.

\bibitem[{Bowman et~al.(2015)Bowman, Angeli, Potts, and
  Manning}]{bowman2015large}
Bowman, S.~R.; Angeli, G.; Potts, C.; and Manning, C.~D. 2015.
\newblock A large annotated corpus for learning natural language inference.
\newblock \emph{arXiv preprint arXiv:1508.05326}.

\bibitem[{Brown et~al.(2020)Brown, Mann, Ryder, Subbiah, Kaplan, Dhariwal,
  Neelakantan, Shyam, Sastry, Askell et~al.}]{brown2020language}
Brown, T.; Mann, B.; Ryder, N.; Subbiah, M.; Kaplan, J.~D.; Dhariwal, P.;
  Neelakantan, A.; Shyam, P.; Sastry, G.; Askell, A.; et~al. 2020.
\newblock Language models are few-shot learners.
\newblock \emph{Advances in neural information processing systems}, 33:
  1877--1901.

\bibitem[{Chen et~al.(2023)Chen, Zhang, Han, Chen, Shi, Xu, and
  Xu}]{chen2023vlp}
Chen, F.-L.; Zhang, D.-Z.; Han, M.-L.; Chen, X.-Y.; Shi, J.; Xu, S.; and Xu, B.
  2023.
\newblock Vlp: A survey on vision-language pre-training.
\newblock \emph{Machine Intelligence Research}, 20(1): 38--56.

\bibitem[{Chen et~al.(2020)Chen, Li, Yu, El~Kholy, Ahmed, Gan, Cheng, and
  Liu}]{chen2020uniter}
Chen, Y.-C.; Li, L.; Yu, L.; El~Kholy, A.; Ahmed, F.; Gan, Z.; Cheng, Y.; and
  Liu, J. 2020.
\newblock Uniter: Universal image-text representation learning.
\newblock In \emph{Computer Vision--ECCV 2020: 16th European Conference,
  Glasgow, UK, August 23--28, 2020, Proceedings, Part XXX}, 104--120. Springer.

\bibitem[{Devlin et~al.(2018)Devlin, Chang, Lee, and
  Toutanova}]{devlin2018bert}
Devlin, J.; Chang, M.-W.; Lee, K.; and Toutanova, K. 2018.
\newblock Bert: Pre-training of deep bidirectional transformers for language
  understanding.
\newblock \emph{arXiv preprint arXiv:1810.04805}.

\bibitem[{Dosovitskiy et~al.(2020)Dosovitskiy, Beyer, Kolesnikov, Weissenborn,
  Zhai, Unterthiner, Dehghani, Minderer, Heigold, Gelly
  et~al.}]{dosovitskiy2020image}
Dosovitskiy, A.; Beyer, L.; Kolesnikov, A.; Weissenborn, D.; Zhai, X.;
  Unterthiner, T.; Dehghani, M.; Minderer, M.; Heigold, G.; Gelly, S.; et~al.
  2020.
\newblock An image is worth 16x16 words: Transformers for image recognition at
  scale.
\newblock \emph{arXiv preprint arXiv:2010.11929}.

\bibitem[{Dou et~al.(2022)Dou, Xu, Gan, Wang, Wang, Wang, Zhu, Zhang, Yuan,
  Peng et~al.}]{dou2022empirical}
Dou, Z.-Y.; Xu, Y.; Gan, Z.; Wang, J.; Wang, S.; Wang, L.; Zhu, C.; Zhang, P.;
  Yuan, L.; Peng, N.; et~al. 2022.
\newblock An empirical study of training end-to-end vision-and-language
  transformers.
\newblock In \emph{Proceedings of the IEEE/CVF Conference on Computer Vision
  and Pattern Recognition}, 18166--18176.

\bibitem[{Goyal et~al.(2017)Goyal, Khot, Summers-Stay, Batra, and
  Parikh}]{goyal2017making}
Goyal, Y.; Khot, T.; Summers-Stay, D.; Batra, D.; and Parikh, D. 2017.
\newblock Making the v in vqa matter: Elevating the role of image understanding
  in visual question answering.
\newblock In \emph{Proceedings of the IEEE conference on computer vision and
  pattern recognition}, 6904--6913.

\bibitem[{Gu et~al.(2021)Gu, Han, Liu, and Huang}]{gu2021ppt}
Gu, Y.; Han, X.; Liu, Z.; and Huang, M. 2021.
\newblock Ppt: Pre-trained prompt tuning for few-shot learning.
\newblock \emph{arXiv preprint arXiv:2109.04332}.

\bibitem[{Han et~al.(2022)Han, Zhao, Ding, Liu, and Sun}]{han2022ptr}
Han, X.; Zhao, W.; Ding, N.; Liu, Z.; and Sun, M. 2022.
\newblock Ptr: Prompt tuning with rules for text classification.
\newblock \emph{AI Open}, 3: 182--192.

\bibitem[{He et~al.(2016)He, Zhang, Ren, and Sun}]{he2016deep}
He, K.; Zhang, X.; Ren, S.; and Sun, J. 2016.
\newblock Deep residual learning for image recognition.
\newblock In \emph{Proceedings of the IEEE conference on computer vision and
  pattern recognition}, 770--778.

\bibitem[{Huang et~al.(2021)Huang, Zeng, Huang, Liu, Fu, and
  Fu}]{huang2021seeing}
Huang, Z.; Zeng, Z.; Huang, Y.; Liu, B.; Fu, D.; and Fu, J. 2021.
\newblock Seeing out of the box: End-to-end pre-training for vision-language
  representation learning.
\newblock In \emph{Proceedings of the IEEE/CVF Conference on Computer Vision
  and Pattern Recognition}, 12976--12985.

\bibitem[{Huang et~al.(2020)Huang, Zeng, Liu, Fu, and Fu}]{huang2020pixel}
Huang, Z.; Zeng, Z.; Liu, B.; Fu, D.; and Fu, J. 2020.
\newblock Pixel-bert: Aligning image pixels with text by deep multi-modal
  transformers.
\newblock \emph{arXiv preprint arXiv:2004.00849}.

\bibitem[{Hudson and Manning(2019)}]{hudson2019gqa}
Hudson, D.~A.; and Manning, C.~D. 2019.
\newblock Gqa: A new dataset for real-world visual reasoning and compositional
  question answering.
\newblock In \emph{Proceedings of the IEEE/CVF conference on computer vision
  and pattern recognition}, 6700--6709.

\bibitem[{Jin et~al.(2021)Jin, Cheng, Shen, Chen, and Ren}]{jin2021good}
Jin, W.; Cheng, Y.; Shen, Y.; Chen, W.; and Ren, X. 2021.
\newblock A good prompt is worth millions of parameters: Low-resource
  prompt-based learning for vision-language models.
\newblock \emph{arXiv preprint arXiv:2110.08484}.

\bibitem[{Kim, Son, and Kim(2021)}]{kim2021vilt}
Kim, W.; Son, B.; and Kim, I. 2021.
\newblock Vilt: Vision-and-language transformer without convolution or region
  supervision.
\newblock In \emph{International Conference on Machine Learning}, 5583--5594.
  PMLR.

\bibitem[{Krishna et~al.(2017)Krishna, Zhu, Groth, Johnson, Hata, Kravitz,
  Chen, Kalantidis, Li, Shamma et~al.}]{krishna2017visual}
Krishna, R.; Zhu, Y.; Groth, O.; Johnson, J.; Hata, K.; Kravitz, J.; Chen, S.;
  Kalantidis, Y.; Li, L.-J.; Shamma, D.~A.; et~al. 2017.
\newblock Visual genome: Connecting language and vision using crowdsourced
  dense image annotations.
\newblock \emph{International journal of computer vision}, 123: 32--73.

\bibitem[{Lester, Al-Rfou, and Constant(2021)}]{lester2021power}
Lester, B.; Al-Rfou, R.; and Constant, N. 2021.
\newblock The power of scale for parameter-efficient prompt tuning.
\newblock \emph{arXiv preprint arXiv:2104.08691}.

\bibitem[{Lewis et~al.(2019)Lewis, Liu, Goyal, Ghazvininejad, Mohamed, Levy,
  Stoyanov, and Zettlemoyer}]{lewis2019bart}
Lewis, M.; Liu, Y.; Goyal, N.; Ghazvininejad, M.; Mohamed, A.; Levy, O.;
  Stoyanov, V.; and Zettlemoyer, L. 2019.
\newblock Bart: Denoising sequence-to-sequence pre-training for natural
  language generation, translation, and comprehension.
\newblock \emph{arXiv preprint arXiv:1910.13461}.

\bibitem[{Li et~al.(2023)Li, Li, Savarese, and Hoi}]{li2023blip}
Li, J.; Li, D.; Savarese, S.; and Hoi, S. 2023.
\newblock Blip-2: Bootstrapping language-image pre-training with frozen image
  encoders and large language models.
\newblock \emph{arXiv preprint arXiv:2301.12597}.

\bibitem[{Li et~al.(2019)Li, Yatskar, Yin, Hsieh, and Chang}]{li2019visualbert}
Li, L.~H.; Yatskar, M.; Yin, D.; Hsieh, C.-J.; and Chang, K.-W. 2019.
\newblock Visualbert: A simple and performant baseline for vision and language.
\newblock \emph{arXiv preprint arXiv:1908.03557}.

\bibitem[{Li et~al.(2020)Li, Yin, Li, Zhang, Hu, Zhang, Wang, Hu, Dong, Wei
  et~al.}]{li2020oscar}
Li, X.; Yin, X.; Li, C.; Zhang, P.; Hu, X.; Zhang, L.; Wang, L.; Hu, H.; Dong,
  L.; Wei, F.; et~al. 2020.
\newblock Oscar: Object-semantics aligned pre-training for vision-language
  tasks.
\newblock In \emph{Computer Vision--ECCV 2020: 16th European Conference,
  Glasgow, UK, August 23--28, 2020, Proceedings, Part XXX 16}, 121--137.
  Springer.

\bibitem[{Liu et~al.(2023{\natexlab{a}})Liu, Li, Wu, and Lee}]{liu2023visual}
Liu, H.; Li, C.; Wu, Q.; and Lee, Y.~J. 2023{\natexlab{a}}.
\newblock Visual instruction tuning.
\newblock \emph{arXiv preprint arXiv:2304.08485}.

\bibitem[{Liu et~al.(2023{\natexlab{b}})Liu, Yuan, Fu, Jiang, Hayashi, and
  Neubig}]{liu2023pre}
Liu, P.; Yuan, W.; Fu, J.; Jiang, Z.; Hayashi, H.; and Neubig, G.
  2023{\natexlab{b}}.
\newblock Pre-train, prompt, and predict: A systematic survey of prompting
  methods in natural language processing.
\newblock \emph{ACM Computing Surveys}, 55(9): 1--35.

\bibitem[{Liu et~al.(2021)Liu, Ji, Fu, Tam, Du, Yang, and Tang}]{liu2021p}
Liu, X.; Ji, K.; Fu, Y.; Tam, W.~L.; Du, Z.; Yang, Z.; and Tang, J. 2021.
\newblock P-tuning v2: Prompt tuning can be comparable to fine-tuning
  universally across scales and tasks.
\newblock \emph{arXiv preprint arXiv:2110.07602}.

\bibitem[{Liu et~al.(2019)Liu, Ott, Goyal, Du, Joshi, Chen, Levy, Lewis,
  Zettlemoyer, and Stoyanov}]{liu2019roberta}
Liu, Y.; Ott, M.; Goyal, N.; Du, J.; Joshi, M.; Chen, D.; Levy, O.; Lewis, M.;
  Zettlemoyer, L.; and Stoyanov, V. 2019.
\newblock Roberta: A robustly optimized bert pretraining approach.
\newblock \emph{arXiv preprint arXiv:1907.11692}.

\bibitem[{Loshchilov and Hutter(2017)}]{loshchilov2017decoupled}
Loshchilov, I.; and Hutter, F. 2017.
\newblock Decoupled weight decay regularization.
\newblock \emph{arXiv preprint arXiv:1711.05101}.

\bibitem[{Lu et~al.(2019)Lu, Batra, Parikh, and Lee}]{lu2019vilbert}
Lu, J.; Batra, D.; Parikh, D.; and Lee, S. 2019.
\newblock Vilbert: Pretraining task-agnostic visiolinguistic representations
  for vision-and-language tasks.
\newblock \emph{Advances in neural information processing systems}, 32.

\bibitem[{Lu et~al.(2022)Lu, Mishra, Xia, Qiu, Chang, Zhu, Tafjord, Clark, and
  Kalyan}]{lu2022learn}
Lu, P.; Mishra, S.; Xia, T.; Qiu, L.; Chang, K.-W.; Zhu, S.-C.; Tafjord, O.;
  Clark, P.; and Kalyan, A. 2022.
\newblock Learn to explain: Multimodal reasoning via thought chains for science
  question answering.
\newblock \emph{Advances in Neural Information Processing Systems}, 35:
  2507--2521.

\bibitem[{Luo et~al.(2023)Luo, Zhou, Ren, Chen, Sun, and Ji}]{luo2023cheap}
Luo, G.; Zhou, Y.; Ren, T.; Chen, S.; Sun, X.; and Ji, R. 2023.
\newblock Cheap and Quick: Efficient Vision-Language Instruction Tuning for
  Large Language Models.
\newblock \emph{arXiv preprint arXiv:2305.15023}.

\bibitem[{OpenAI(2022)}]{chatgpt}
OpenAI. 2022.
\newblock ChatGPT.
\newblock Online.

\bibitem[{Plummer et~al.(2015)Plummer, Wang, Cervantes, Caicedo, Hockenmaier,
  and Lazebnik}]{plummer2015flickr30k}
Plummer, B.~A.; Wang, L.; Cervantes, C.~M.; Caicedo, J.~C.; Hockenmaier, J.;
  and Lazebnik, S. 2015.
\newblock Flickr30k entities: Collecting region-to-phrase correspondences for
  richer image-to-sentence models.
\newblock In \emph{Proceedings of the IEEE international conference on computer
  vision}, 2641--2649.

\bibitem[{Radford et~al.(2021)Radford, Kim, Hallacy, Ramesh, Goh, Agarwal,
  Sastry, Askell, Mishkin, Clark et~al.}]{radford2021learning}
Radford, A.; Kim, J.~W.; Hallacy, C.; Ramesh, A.; Goh, G.; Agarwal, S.; Sastry,
  G.; Askell, A.; Mishkin, P.; Clark, J.; et~al. 2021.
\newblock Learning transferable visual models from natural language
  supervision.
\newblock In \emph{International conference on machine learning}, 8748--8763.
  PMLR.

\bibitem[{Raffel et~al.(2020)Raffel, Shazeer, Roberts, Lee, Narang, Matena,
  Zhou, Li, and Liu}]{raffel2020exploring}
Raffel, C.; Shazeer, N.; Roberts, A.; Lee, K.; Narang, S.; Matena, M.; Zhou,
  Y.; Li, W.; and Liu, P.~J. 2020.
\newblock Exploring the limits of transfer learning with a unified text-to-text
  transformer.
\newblock \emph{The Journal of Machine Learning Research}, 21(1): 5485--5551.

\bibitem[{Rao et~al.(2021)Rao, Zhao, Liu, Lu, Zhou, and
  Hsieh}]{rao2021dynamicvit}
Rao, Y.; Zhao, W.; Liu, B.; Lu, J.; Zhou, J.; and Hsieh, C.-J. 2021.
\newblock Dynamicvit: Efficient vision transformers with dynamic token
  sparsification.
\newblock \emph{Advances in neural information processing systems}, 34:
  13937--13949.

\bibitem[{Ren, Kiros, and Zemel(2015)}]{ren2015exploring}
Ren, M.; Kiros, R.; and Zemel, R. 2015.
\newblock Exploring models and data for image question answering.
\newblock \emph{Advances in neural information processing systems}, 28.

\bibitem[{Ren et~al.(2015)Ren, He, Girshick, and Sun}]{ren2015faster}
Ren, S.; He, K.; Girshick, R.; and Sun, J. 2015.
\newblock Faster r-cnn: Towards real-time object detection with region proposal
  networks.
\newblock \emph{Advances in neural information processing systems}, 28.

\bibitem[{Ryoo et~al.(2021)Ryoo, Piergiovanni, Arnab, Dehghani, and
  Angelova}]{ryoo2021tokenlearner}
Ryoo, M.~S.; Piergiovanni, A.; Arnab, A.; Dehghani, M.; and Angelova, A. 2021.
\newblock Tokenlearner: What can 8 learned tokens do for images and videos?
\newblock \emph{arXiv preprint arXiv:2106.11297}.

\bibitem[{Shen et~al.(2021)Shen, Li, Tan, Bansal, Rohrbach, Chang, Yao, and
  Keutzer}]{shen2021much}
Shen, S.; Li, L.~H.; Tan, H.; Bansal, M.; Rohrbach, A.; Chang, K.-W.; Yao, Z.;
  and Keutzer, K. 2021.
\newblock How much can clip benefit vision-and-language tasks?
\newblock \emph{arXiv preprint arXiv:2107.06383}.

\bibitem[{Sung, Cho, and Bansal(2022)}]{sung2022vl}
Sung, Y.-L.; Cho, J.; and Bansal, M. 2022.
\newblock Vl-adapter: Parameter-efficient transfer learning for
  vision-and-language tasks.
\newblock In \emph{Proceedings of the IEEE/CVF Conference on Computer Vision
  and Pattern Recognition}, 5227--5237.

\bibitem[{Tan and Bansal(2019)}]{tan2019lxmert}
Tan, H.; and Bansal, M. 2019.
\newblock Lxmert: Learning cross-modality encoder representations from
  transformers.
\newblock \emph{arXiv preprint arXiv:1908.07490}.

\bibitem[{Touvron et~al.(2023)Touvron, Lavril, Izacard, Martinet, Lachaux,
  Lacroix, Rozi{\`e}re, Goyal, Hambro, Azhar et~al.}]{touvron2023llama}
Touvron, H.; Lavril, T.; Izacard, G.; Martinet, X.; Lachaux, M.-A.; Lacroix,
  T.; Rozi{\`e}re, B.; Goyal, N.; Hambro, E.; Azhar, F.; et~al. 2023.
\newblock Llama: Open and efficient foundation language models.
\newblock \emph{arXiv preprint arXiv:2302.13971}.

\bibitem[{Tsimpoukelli et~al.(2021)Tsimpoukelli, Menick, Cabi, Eslami, Vinyals,
  and Hill}]{tsimpoukelli2021multimodal}
Tsimpoukelli, M.; Menick, J.~L.; Cabi, S.; Eslami, S.; Vinyals, O.; and Hill,
  F. 2021.
\newblock Multimodal few-shot learning with frozen language models.
\newblock \emph{Advances in Neural Information Processing Systems}, 34:
  200--212.

\bibitem[{Wang et~al.(2021)Wang, Yu, Yu, Dai, Tsvetkov, and
  Cao}]{wang2021simvlm}
Wang, Z.; Yu, J.; Yu, A.~W.; Dai, Z.; Tsvetkov, Y.; and Cao, Y. 2021.
\newblock Simvlm: Simple visual language model pretraining with weak
  supervision.
\newblock \emph{arXiv preprint arXiv:2108.10904}.

\bibitem[{Wei et~al.(2021)Wei, Bosma, Zhao, Guu, Yu, Lester, Du, Dai, and
  Le}]{wei2021finetuned}
Wei, J.; Bosma, M.; Zhao, V.~Y.; Guu, K.; Yu, A.~W.; Lester, B.; Du, N.; Dai,
  A.~M.; and Le, Q.~V. 2021.
\newblock Finetuned language models are zero-shot learners.
\newblock \emph{arXiv preprint arXiv:2109.01652}.

\bibitem[{Xie et~al.(2019)Xie, Lai, Doran, and Kadav}]{xie2019visual}
Xie, N.; Lai, F.; Doran, D.; and Kadav, A. 2019.
\newblock Visual entailment: A novel task for fine-grained image understanding.
\newblock \emph{arXiv preprint arXiv:1901.06706}.

\bibitem[{Yan et~al.(2021)Yan, Xu, Li, Bi, Tian, Gui, and Wang}]{yan2021grid}
Yan, M.; Xu, H.; Li, C.; Bi, B.; Tian, J.; Gui, M.; and Wang, W. 2021.
\newblock Grid-vlp: Revisiting grid features for vision-language pre-training.
\newblock \emph{arXiv preprint arXiv:2108.09479}.

\bibitem[{Yang et~al.(2016)Yang, He, Gao, Deng, and Smola}]{yang2016stacked}
Yang, Z.; He, X.; Gao, J.; Deng, L.; and Smola, A. 2016.
\newblock Stacked attention networks for image question answering.
\newblock In \emph{Proceedings of the IEEE conference on computer vision and
  pattern recognition}, 21--29.

\bibitem[{Yu et~al.(2019)Yu, Yu, Cui, Tao, and Tian}]{yu2019deep}
Yu, Z.; Yu, J.; Cui, Y.; Tao, D.; and Tian, Q. 2019.
\newblock Deep modular co-attention networks for visual question answering.
\newblock In \emph{Proceedings of the IEEE/CVF conference on computer vision
  and pattern recognition}, 6281--6290.

\bibitem[{Zhou et~al.(2022{\natexlab{a}})Zhou, Yang, Loy, and
  Liu}]{zhou2022conditional}
Zhou, K.; Yang, J.; Loy, C.~C.; and Liu, Z. 2022{\natexlab{a}}.
\newblock Conditional prompt learning for vision-language models.
\newblock In \emph{Proceedings of the IEEE/CVF Conference on Computer Vision
  and Pattern Recognition}, 16816--16825.

\bibitem[{Zhou et~al.(2022{\natexlab{b}})Zhou, Yang, Loy, and
  Liu}]{zhou2022learning}
Zhou, K.; Yang, J.; Loy, C.~C.; and Liu, Z. 2022{\natexlab{b}}.
\newblock Learning to prompt for vision-language models.
\newblock \emph{International Journal of Computer Vision}, 130(9): 2337--2348.

\bibitem[{Zhou et~al.(2019)Zhou, Ji, Su, Sun, and Chen}]{zhou2019dynamic}
Zhou, Y.; Ji, R.; Su, J.; Sun, X.; and Chen, W. 2019.
\newblock Dynamic capsule attention for visual question answering.
\newblock In \emph{Proceedings of the AAAI conference on artificial
  intelligence}, volume~33, 9324--9331.

\bibitem[{Zhou et~al.(2020)Zhou, Ji, Sun, Luo, Hong, Su, Ding, and
  Shao}]{zhou2020k}
Zhou, Y.; Ji, R.; Sun, X.; Luo, G.; Hong, X.; Su, J.; Ding, X.; and Shao, L.
  2020.
\newblock K-armed bandit based multi-modal network architecture search for
  visual question answering.
\newblock In \emph{Proceedings of the 28th ACM international conference on
  multimedia}, 1245--1254.

\bibitem[{Zhou et~al.(2021)Zhou, Ren, Zhu, Sun, Liu, Ding, Xu, and
  Ji}]{zhou2021trar}
Zhou, Y.; Ren, T.; Zhu, C.; Sun, X.; Liu, J.; Ding, X.; Xu, M.; and Ji, R.
  2021.
\newblock Trar: Routing the attention spans in transformer for visual question
  answering.
\newblock In \emph{Proceedings of the IEEE/CVF International Conference on
  Computer Vision}, 2074--2084.

\end{thebibliography}

\end{document}